\documentclass{svproc}

\usepackage{graphicx}%
\usepackage{multirow}%
\usepackage{amsmath,amssymb,amsfonts}%
\usepackage{mathrsfs}%
\usepackage[title]{appendix}%
\usepackage{xcolor}%
\usepackage{textcomp}%
\usepackage{manyfoot}%
\usepackage{booktabs}%
\usepackage{algorithm}[lined, box]

\usepackage{algorithmicx}%
\usepackage{algpseudocode}%
\usepackage{listings}%
\usepackage{url}

\usepackage{adjustbox}
\usepackage{cleveref}
\usepackage{verbatim}
  
\usepackage{color, url}
\newcounter{noteAEctr} 
\newcommand{\AEE}[1]{ \textbf{\textcolor{purple}{{{AE:  }}#1}} 
\newcommand{\AEEr}[1]{ {\textcolor{black}{{{ }}#1}}} \addtocounter{noteAEctr}{1} }
\newcounter{noteTonyctr} 
\newcommand{\tony}[1]{ \textbf{\textcolor{black}{{{Tony: \#\arabic{noteTonyctr}: }}#1}}
\addtocounter{noteTonyctr}{1} }

\newcounter{noteGesinectr} 
\newcommand{\gr}[1]{ \textbf{\textcolor{black}{{{GR: \#\arabic{noteGesinectr}: }}#1}}
\addtocounter{noteGesinectr}{1} }
\newcommand{\gdr}[1]{\textcolor{black}{#1}}

\newcounter{noteStratisctr} 

\newcommand{\stratis}[1]{\textcolor{black}{#1}}

\newcommand{\mc}[1]{\textcolor{black}{#1}}

\newcommand{\lgg}{L2G2G\ }

\newcommand{\AEEr}[1]{\textcolor{black}{ AE: #1}}

\begin{document}
\mainmatter              
\title{L2G2G: a Scalable Local-to-Global Network Embedding with Graph Autoencoders}
\titlerunning{Local-2-GAE-2-Global}  
%
\author{Ruikang Ouyang\inst{1} \and Andrew Elliott\inst{5,6}
Stratis Limnios\inst{5} \and Mihai Cucuringu\inst{2,3, 4, 5}\and Gesine Reinert \inst{2,5}}
\authorrunning{R. Ouyang et al.} 

\institute{Department of Engineering, University of Cambridge, Cambridge, UK 
\and Department of Statistics, University of Oxford, Oxford, UK
\and Mathematical Institute, University of Oxford, Oxford, UK
\and Oxford-Man Institute of Quantitative Finance, University of Oxford, Oxford, UK
\and
The Alan Turing Institute, London, UK
\and School of Mathematics and Statistics, University of Glasgow, Glasgow, UK
}

\maketitle              
\begin{abstract}
\gdr{For analysing real-world networks, graph representation learning is a popular tool. These methods, such as a graph autoencoder (GAE), typically rely on low-dimensional representations, also called \textit{embeddings}, which are obtained through minimising a loss function; these embeddings are used with a decoder 
for  downstream tasks such as node classification and edge prediction.}
 While GAEs tend to be fairly accurate, they suffer from scalability issues. 
 For 
improved speed, a  
 Local2Global approach, which combines graph patch embeddings based on eigenvector synchronisation, \mc{was} shown to be fast and achieve good accuracy. 
Here we propose L2G2G, a Local2Global method which improves {GAE} accuracy without sacrificing scalability. 
This  improvement is achieved by dynamically synchronising  the latent node representations, while training the GAEs. It also benefits from the decoder computing an \gdr{only} local patch loss. 
 Hence, aligning the local embeddings in each epoch utilises more information from the graph than a single post-training alignment does, while maintaining scalability. We illustrate on synthetic benchmarks, as well as real-world examples, that L2G2G achieves higher accuracy than the standard Local2Global approach and scales efficiently on the larger data sets. We find that for large and dense networks, it even outperforms the 
slow, but assumed more accurate, GAEs.

\keywords{Graph Autoencoder, Local2Global, 
Node Embedding, \mc{Group Synchronisation}}
\end{abstract}
\section{Introduction}
\stratis{Graph representation learning has been a core component in graph based real world applications, \gdr{for an introduction see \cite{hamilton2020graph}.} 
\gdr{A}s graphs have become ubiquitous in a wide array of applications, 
low-dimensional representations \gdr{are needed} to tackle the curse of dimensionality inherited by the graph structure. 
}
In practice,
{l}ow-dimensional node embeddings are {used as} efficient representations to address
tasks such as graph clustering \cite{Graph-clustering-with-GNN},
node classification \cite{bayer2022label}, 
and link prediction \cite{link-prediction-on-gnn}, 
or 
to protect private data in federated learning settings \cite{FL+GNN1,he2021fedgraphnn}.

\stratis{Graph Autoencoders (GAEs) \cite{simonovsky2018graphvae,vgae} emerged as a powerful \gdr{Graph Neural Network (GNN) \cite{bruna2013spectral}} tool to produce \gdr{such} node representations}. 
\gdr{GAEs} 
adapt autoencoders 
and variational autoencoders 
\cite{pmlr-v27-baldi12a,vgae} to graph structure data using a Graph Convolutional Neural Network (GCN) \cite{gcn} as the encoder and 
\gdr{for} 
node embeddings.
Although a GAE can achieve 
\gdr{high} accuracy in graph reconstruction, it suffers from 
{a} high computational cost. 
 Several solutions \mc{for reducing computational workload} have been proposed. Linear models, like PPRGo~\cite{pprgo} and SGC~\cite{chen2020simple}, remove the non-linearities between layers.  
  Layer-wise sampling methods  such as GraphSage~\cite{hamilton2017inductive}, FastGCN~\cite{chen2018fastgcn} and LADIES~\cite{LADIES} 
 \gdr{sample} a subset of neighbors of the nodes in each layer, \gdr{while} 
 subgraph-sampling based methods  such as GraphSaint~\cite{graphsaint} and Cluster-GCN~\cite{cluster-gcn} 
 \gdr{carry out} message passing \gdr{only through a sampled subgraph}. 

In this paper, we use FastGAE~\cite{fastgae} as a starting point, 
which computes approximate reconstruction losses by evaluating their values only from \mc{suitably} selected random subgraphs of the original graph. 
\gdr{W}hile FastGAE reduces the computational cost of a traditional GAE, its overall performance can be substantially inferior to a GAE when the sample used to approximate the loss is not large enough. \stratis{For improved performance, the general Local2Global (L2G) framework by  \cite{local2global} 
leverages the eigenvector synchronization of  \cite{es1,es2},
to align independently created embeddings in order to produce a globally consistent structure (\gdr{here we employ GAEs for embeddings and denote the resulting method by GAE$+$L$2$G}). However, 
this architecture is data inefficient and suffers from a loss of performance in downstream tasks since \gdr{it learns} 
multiple separate GAEs. Moreover  aggregating the local embeddings after the training process might lead to a loss of  
useful information learned during training.}

\begin{figure}[t]
    \centering
    \includegraphics[width=0.95\linewidth]{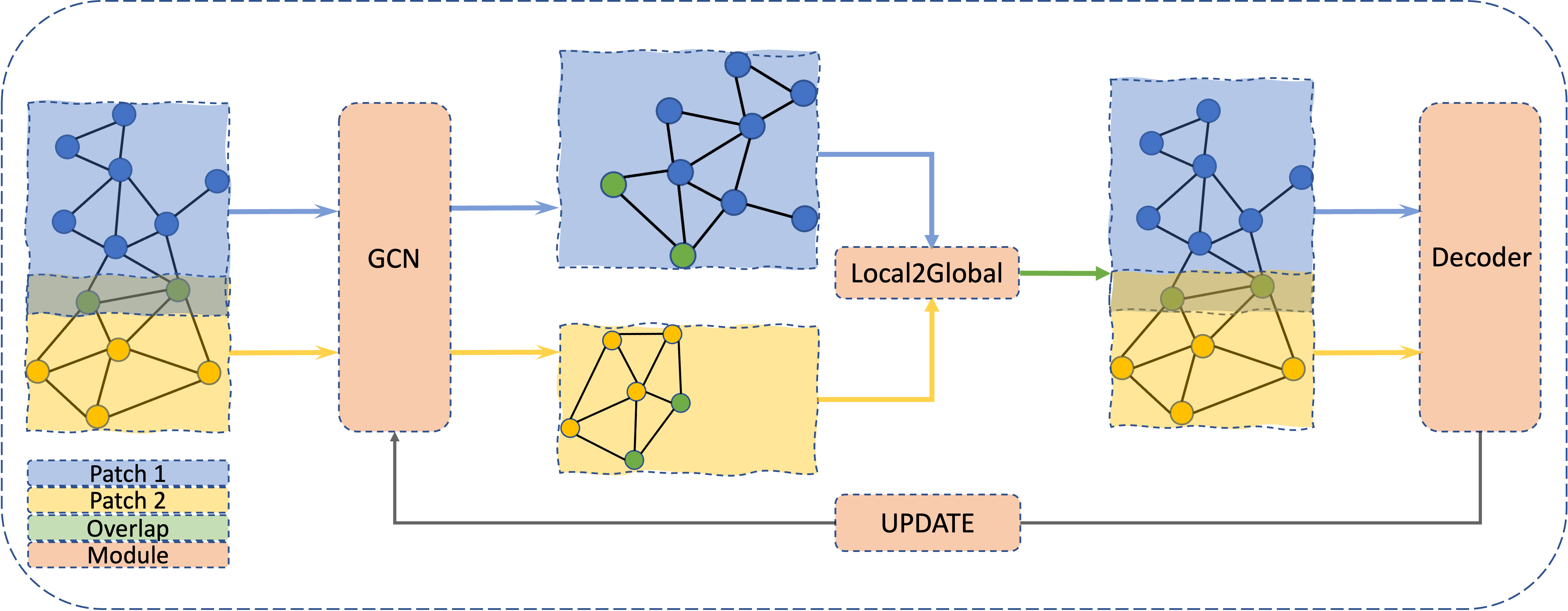}
    \caption{L2G2L pipeline 
     {for two patches.} %
    {T}he two patches  {are} in blue and yellow, 
    the overlapping nodes between them in green.
    Separate node embeddings for each patch are obtained via a single GCN. The decoder aligns the embeddings using the Local2Global synchronisation algorithm to yield a global embedding and then uses a standard sigmoid function. The GCN is then iteratively optimised using the training loss.}
    \label{fig:pipeline}
\end{figure}

\gdr{Instead,} 
we propose the Local to GAE to Global (L2G2G) model, which optimizes the Local2Global aligned embeddings directly, reducing the amount of information loss and allowing us to train a single global modified GAE. 
\gdr{T}his structure 
leverage\gdr{s} the scalable approach of FastGAE by only considering small sections of the graph when updating the weights. Figure \ref{fig:pipeline} \gdr{shows the model} 
pipeline.  
Our main contributions are:
\\
1. We introduce L2G2G as a new fast network embedding method. 
\\
2. \stratis{We provide 
a theoretical complexity analysis for GAE$+$L$2$G and an experimental comparison of the runtimes showing that \gdr{the} 
runtime sacrifice for performance \gdr{in L2G2G} is minimal.}
%
\\
3. We test 
L2G2G and the baseline methods on real and synthetic data sets, demonstrating 
that L2G2G can  boost the performance of GAE$+$L$2$G,  
especially on  medium scale data sets, 
while achieving comparable training speed.  
%

\stratis{The paper is 
structured as follows. \gdr{Section \ref{sec:pre}} 
introduces 
notation and discusses GAEs 
\gdr{and} the Local2Global framework by \cite{local2global}, including GAE$+$L$2$G. Section~\ref{sec:l2g2g} presents our method, L2G2G, as well as a time complexity analysis, comparing it to GAE, FastGAE,and GAE$+$L$2$G. Section~\ref{sec:experiment} provides experimental results on synthetic and real data sets, on networks of up to about $700,000$ nodes. In Section~\ref{sec:conclusion} we discuss the results and indicate directions for future work.}
The code is  available at {\url{https://github.com/tonyauyeung/Local2GAE2Global}.}

\section{Preliminaries}\label{sec:pre}

\paragraph{Notations:}
{An}
undirected \gdr{attributed} graph $G=(V, E, X)$
consists of 
a set of nodes $V$ of size $N$, a set of unweighted, undirected edges $E$ of size $M$, and a 
{$N \times F$ matrix $X$} of real-valued node attributes (features). \gdr{The edge set is also represented by the $N \times N$ adjacency matrix $A$.} \stratis{Moreover, based on the L2G framework, we define a patch $P$ to be a subgraph of $G$ which is induced by a subset of the node set $V$; hence a patch $P_i$ with the feature matrix corresponding to its nodes is denoted as 
$(V^{(i)},E^{(i)},X^{(i)})$ 
Node embeddings are denoted as 
a $N \times e$ matrix $Z$}, where $e$ is the embedding size {and $\sigma$ is the sigmoid function.} 


\paragraph{Graph Autoencoders {(GAEs)}:}

Given a graph  {$G=(V, E, X)$},
 {a GCN is used to obtain} 
an $N \times e$ embedding matrix
\gdr{$Z=GCN(X, A)$},  and \gdr{a sparse approximation of the adjacency matrix through} 
 %
 $\hat{A}=\sigma(ZZ^T).$
 The GAE \gdr{obtains the embedding through minimising} 
 the cross-entropy reconstruction loss,   
$L_{GAE}(\hat{A}, A) =Loss(\hat{A}, A)
: =
 -\sum_{i,j=1}^{N}A_{ij}log\hat{A}_{ij} 
$
\gdr{with respect to the parameters of the GCN; 
this minimisation is also called {\it training}. Here a recursive method called {\it message passing} is used. The {\it decoder} then 
computes an inner product between each pair of node embeddings in the graph as proxy for the edge probabilities.}
Even though GAEs outperform traditional node embedding methods, such as spectral clustering~\cite{sc} and DeepWalk~\cite{dw}, they usually scale poorly 
to large graphs. 
This is due to 
having to visit all the neighbors of a node recursively during the message passing phase in the encoding GCN, {and} 
the decoder %
{scaling} 
as $O(N^2)$ in complexity.
We highlight two approaches for improving scalability: 

{1. FastGAE}
\cite{fastgae}: 
\stratis{This model addresses the scalability issues by reconstructing the adjacency matrix of a sampled subgraph. This is achieved by evaluating an approximate reconstruction loss ($Loss_{approx}$) for every subgraph and aggregating them in one loss to be optimized by the model.
This sampling procedure reduces the computation complexity of decoding during each training epoch 
from $O(N^2)$ to $O(N_S^2)$, where $N_S$ is the number of nodes in the subgraph.}

{2. Local2Global {(L2G)}}
\cite{local2global}:  
\stratis{This framework is a generic method to }
 align embeddings computed on
different parts of the graph (potentially on different machines \mc{and by different entities with different privacy constraints}) into a single 
global embedding, \stratis{regardless of the embedding method, }
as follows. 
Suppose that $P_1,...,P_k$ are $k$ patches, which pairwise overlap on at least $d$ nodes and at most $l$ nodes.  It is assumed that the graph union of all patches gives the initial graph $G$.  The pattern of overlapping patches 
is captured in a so-called \textit{patch graph}, denoted $G_P = (V_P, E_P)$, whose node set $V_P = \{P_1,\dots,P_k\}$ denote the patches. 
An edge between two nodes in $G_P$ \stratis{indicates that there is an overlap of at least $d$ nodes in the initial graph $G$ between those two patches.
Then, for each patch $P_i$ a node embedding matrix $Z_i$ is computed} 
using an embedding method of choice. \gdr{When the embedding is obtained through a GAE, we refer to the method as GAE+L2G.}
Local2Global \gdr{then} leverages the overlap
of the patches to compute an optimal alignment based on a set of 
affine transforms which synchronizes all the local patch embeddings into a single and globally consistent embedding, \gdr{as follows.} 
First, we \gdr{estimate} 
rotation matrices $\hat{S}_j \gdr{\in \mathbb{R}^{F \times F}}, 
j
=1,\ldots, k$, one for each patch. \gdr{With}
  $M_{ij}=\sum_{u\in P_i\cap P_j}X^{(i)}_u(X^{(i)}_u)^T $
    we first estimate the rotations between each pair of overlapping 
patches 
$(P_i, P_j)\in E_p$ by 
 $R_{ij}=M_{ij}(M_{ij}^TM_{ij})^{-1/2}.$ 
Next we build
$
    \Tilde{R}_{ij}={w_{ij}R_{ij}}/{\sum_j 
    |{V(P_i)\cap V(P_j)}|}  
    $ 
to 
approximately solve the eigen problem  $S=\Tilde{R}S$, \gdr{obtaining}  $\hat{S}=[\hat{S}_1,...\hat{S}_k]$. 
We also 
\gdr{find} a translation matrix $\hat{T}=[\hat{T}_1, \ldots, \hat{T}_k]$ by  solving 
$
    \hat{T}=\underset{{T \in \mathbb{R}^{k\times F}}}{\arg min}||BT-C||_2^2, 
$
where $B\in \{-1, 1\}^{|E_p| \times k}$ is the incidence matrix of the patch graph with entries $B_{(P_i, P_j), t} = \delta_{it} - \delta_{jt}$, $\delta$  is the Kronecker Delta, and 
$C\in \mathbb{R}^{|E_p| \times F}$ has entries 
$C_{(P_i, P_j)} = 
\sum_{t \in P_i \cap P_j}\big(\hat{Z}^{(i)}_t - \hat{Z}^{(j)}_t\big)/ {|P_i \cap P_j|}$. 
This solution yields the estimated coordinates of all the nodes up to a global rigid transformation.
Next, we \gdr{apply} 
the appropriate rotation transform to each patch individually, 
  $  \hat{Z}^{(j)}=Z^{(j)}\hat{S}_j^T, $
{then a}pply the corresponding translation to each patch (\mc{hence performing translation synchronisation}), and \mc{finally average in order to obtain}  
 the final aligned node embedding 
$
    \Bar{Z}_i={{\sum_j(\hat{Z}_i^{(j)} + \hat{T}_j})} / {|\{j \ :\ i \in P_j\}|}.
    $

\section{Methodology}
\label{sec:l2g2g}

\paragraph{Local-2-GAE-2-Global}
Combining Local2Global and GAEs produces a scalable GAE extension for node embeddings using autoencoders; using separate GAEs for each of the patches 
allows specialization 
to  the unique structure in each
of the patches. Our {Local-2-GAE-2-Global} (L2G2G) framework 
leverages the same divide-and-conquer technique Local2Global capitalises on, but  is designed and adapted to the  traditional GAE pipeline \gdr{to benefit from its accuracy}. 
\stratis{The core idea of \gdr{L2G2G} 
is to evaluate embeddings locally on the patches but 
synchroniz\gdr{ing} the patch embeddings using the L2G framework while training 
\gdr{a} GAE. This leads to 
$k$ GCNs encoding  \gdr{the $k$} patches:}
	$Z_i = GCN(X^{(i)},A^{(i)}), $ {for $i=1, \ldots, k$.}
To account for \stratis{the dynamic update during training and adapt to the local optimization scheme}, 
we modify 
the GAE decoder to adjust the embeddings using the Local2Global framework; \stratis{hence
the patch-wise decoder in L2G2G \gdr{estimates the edge probabilities between nodes in patches $i$ and $j$ by}} 
$\sigma((S_iZ_i+T_i)^T(S_jZ_j+T_j)),$
where $\gdr{S_i=} S_i(Z)$ and $\gdr{T_i=} T_i(Z)$ are {the Local2Global transformations} 
of each of the patch embeddings.

\stratis{In contrast to  GAE+L2G, \gdr{L2G2G} synchronizes the embeddings before the decoder step and also performs synchronizations during the model training, 
thus taking full advantage of the patch graph structure during training.}
\gdr{The} 
cross-entropy losses \mc{of} 
each patch \gdr{are aggregated} to \gdr{give} a global loss function: 
\[
L_{L2G2G} =  \sum_{j=1}^{k}{N_j}L_{GAE}\left(\hat{A}^{(j)}, A^{(j)}
\right) / N . 
\]
Similarly to the FastGAE algorithm, L2G2G reduces computation
by only considering local structure.
However, rather than training the network using only the local information,
L2G2G aggregates the local embeddings to reconstruct the global information, 
thus boosting performance.  

A schematic diagram of L2G2G is shown 
in figure~\ref{fig:pipeline}, and \gdr{pseudo-code for L2G2G is given in algorithm~\ref{alg:Local2GAE2Global}.}
\gdr{A}s computing the alignment step can be 
costly, 
assum\gdr{ing} that the 
Local2Global alignment
would not change too quickly during training, \gdr{we}
 update the {rotation and translation} matrices only every 10 epochs. 

\begin{algorithm}[ht]
\caption{Local-2-GAE-2-Global {(L2G2G)}: An overview}
\label{alg:Local2Global FastGAE}
\begin{algorithmic}
\Require $P_1,...,P_k$, where $P_j=(X^{(j)}, A^{(j)})$
\For{$e$ in $[1,...,T]$}
\For{$j$ in $[1,...k]$}
\State$Z_j \gets GCN(X^{(j)}, A^{(j)})$
\EndFor
\State$\hat{Z}_1,...,\hat{Z}_k \gets Sync(Z_1,...,Z_k)$
\State$L \gets 0$
\For{$j$ in $[1,...,k]$}
\State$\hat{A}_j \gets \sigma(\hat{Z}_j\hat{Z}_j^T)$
\State$L \gets L + \frac{N_j}{N}L_{GAE}(\hat{A}^{(j)}, A^{(j)})$
\EndFor
\State Optimize encoder using $L$ 
\EndFor
\end{algorithmic}
\label{alg:Local2GAE2Global}
\end{algorithm}
\vspace{-0.5cm}
\paragraph{Complexity Analysis}
Following the computations in \cite{scalable-gnn-via-bp} and \cite{local2global}, we derive the complexity of GAE, FastGAE, GAE+{L2G} and L2G2G.
We assume that the number of nodes, edges and features satisfy $N,M,F \ge 1$, and, following
\cite{scalable-gnn-via-bp}, 
that the dimensions of the hidden layers in the GCN are all $F$. Then, the complexity of a $L$-layer GCN  scales like $O(LNF^2 + LMF)$ and that of the inner product decoder scales like $O(N^2F)$. \AEEr{maybe add something here about full decoder?} Thus, for as shown in \cite{scalable-gnn-via-bp}, for $T$ epochs the time complexity of the decoder and the encoder of a GAE scales like $ O({T}(LNF^2 + LMF + N^2F))$.    
In contrast, as stated in \cite{fastgae}, the complexity of per-epoch of FastGAE with a $\sqrt{N}$ down-sampling size is
$ O(LNF^2+LMF+NF)$, and hence for $T$ epochs the {FastGAE} complexity scales like $O(T(LNF^2+LMF+NF)).$

{To simplify the complexity analysis} 
of both Local2Global approaches 
we assume that the overlap size of two overlapping patches in the patch graph 
is fixed to $d \sim F$.
Following \cite{local2global},
finding the rotation matrix $S$ scales like $O(|E_p|dF^2) = 
O(|E_p|F^3)$. 
The translation problem can be solved by a $t$-iteration solver with a complexity per iteration of $O(|E_p|F)$, where $t$ is fixed.
To align the local embeddings, one has to perform  matrix multiplications, which requires $O(N_jF^2)$ computations, where $N_j$ is the number of nodes in the $j^{th}$ patch.
The complexity of finding the rotation matrix 
($O(|E_p|F^3)$)  dominates  the complexity of the computing the translation ($O(|E_p|F)$).
Thus, the complexity of the L2G algorithm with $k$ patches is 
$ O\big(|E_p|F^3 + F^2 \sum_{j=1}^{k}N_j \big). $  

The GAE+L2G algorithm uses a GAE for every patch, and for the $j^{th}$ patch, for $T$ training epochs the GAE  scales like $O(T( L N_j F^2 + L M_j F + N_j^2 F)),$ with $M_j$ 
number of edges in the $j^{th}$ patch. Summing over all patches and ignoring the overlap between patches as lower order term, so that $\sum_j N_j = O(N)$, $\sum_j N_j^2 \approx N^2/k,$ and $\sum_j M_j \approx M,$ the GAE+L2G algorithm scales like 
$  O\left( TF \left(LNF    + LM + 
    {N}/{k} 
    \right)
    +k F^3 
    \right).
    $
{For} the complexity of L2G2G, 
{as} 
L2G2G aligns the local embeddings in each epoch rather than after training, we replace $ k F^3 + NF^2$  with $T( k F^3 + NF^2)$, {and thus the algorithm} 
scales like 
  $  O\Big( T \Big(LNF^2    + LMF + 
    \frac{N^2}{k}F 
    +
    k F^3  
   \Big)
    \Big) .$
{In the PyTorch implementation \gdr{of FastGAE}, 
the reconstruction error is 
\gdr{approximated} by  \gdr{creating the induced subgraph from  sampling $\lfloor \sqrt{N} \rfloor$ proportional to degree}, with an expected number of  at least $O(M/N)$ edges between them. Then, the computation of the decoder 
\gdr{is (at least)} $O(\gdr{M/N})$ instead of $O(N^2)$.} 
\gdr{\Cref{tab:compare-general} summarises the complexity results.} 
\vspace{-0.5cm}
\begin{table*}[ht]
\begin{adjustbox}{width=0.8\linewidth, center}
\begin{tabular}{c l l}
\toprule
Model           & General Time Complexity                                                 & PyTorch implementation  \\ 
\midrule
GAE             & $O\left(TF(LNF + LM + N^2)\right)$                                      & $O\left(TF (LNF + LM+ M)\right)$ \\
FastGAE         & $\ge O\left(TF(LNF + LM + N)\right)$                                        & $O\left(TF(LNF + LM + \gdr{M/N})\right)$ \\
GAE+L2G         & $O\left(TF(L(NF + M) + \frac{N^2}{k})+kF^3\right) $                      &  $O\left(TF(LNF + LM + M)+kF^3\right) $\\
L2G2G           & $O\left(TF(L(NF + M) + \frac{N^2}{k}+kF^3)\right)$                       &  $O\left(TF(LNF + LM + M +kF^3)\right)$
\\
\bottomrule
\end{tabular}
\end{adjustbox}
\caption{\label{tab:compare-general}Complexity comparison in the  general  {and in the sparse case} 
.}
\end{table*}
\vspace{-0.5cm}

\gdr{Thus,} in the \gdr{standard} case,  
increasing the number of patches $k$ reduces 
the complexity of the computation of the GAE decoders.
In the \gdr{PyTorch implementation,} 
if 
$k$ scales linearly with $N$, the expression 
\gdr{is} 
linear in $N$. 
In contrast, when the number of nodes $N$ is not very large, the number of features $F$ becomes more prominent, so that the training speed may not necessarily increase with increasing number of patches. 
\Cref{tab:compare-general} 
shows that L2G2G sacrifices $O\left(TkF^3\right)$ training time to obtain better performance;  \stratis{
with an increase in the number of patches, the training speed gap between L2G2G and GAE+L2G increases linearly.}

\section{Experimental Evaluation} \label{sec:experiment}

\paragraph{Datasets} 
To measure the performance of our method, we compare the ability of 
L2G2G to
\stratis{learn node embeddings for graph reconstruction
against the following benchmark datasets }
Cora ML, 
Cora~\cite{Cora&ML}, Reddit~\cite{graphsaint} and Yelp~\cite{graphsaint}.

In addition, we tested  the performance of L2G2G on four synthetic data sets,  
generated using a Stochastic Block Model (SBM) 
which assigns nodes to blocks; edges are placed independently between nodes in a block with probability $p_{in}$
and between blocks with probability $p_{out}$ \cite{SBM}. \stratis{We \gdr{encode} 
the block membership as node features; \gdr{with $L$ blocks,  $v$ being in block $l$ is encoded as  unit vector $e_l \in \{0,1\}^L$.}} 
\gdr{T}o test the performance across multiple scales
we fix the number of blocks at $100$, and vary the block size, $p_{in}$ and $p_{out}$,  \gdr{as follows:}

1. `SBM-Small' with block size{s} $10^2$ 
and $(p_{in},p_{out})=(0.02,10^{-4})$,

2. 
`SBM-Large-Sparse' with block size{s} $10^3$ 
and $(p_{in},p_{out})=(10^{-3},10^{-4})$,  

3. 
`SBM-Large' with blocks of size{s} $10^3$  and $(p_{in},p_{out})=(0.02,10^{-4})$,  

 4. `SBM-Large-Dense' with block {sizes} $10^3$ and $(p_{in},p_{out})=(0.1,0.002)$. 

\noindent\gdr{\Cref{tab:dataset} gives some summary statistics} of these \gdr{real and synthetic} data sets.
{\small 
{\small 
\begin{table} 
\begin{tabular}{@{}l|cccc|cccc@{}}
\toprule
           & \multicolumn{4}{c|}{Stochastic block model} & \multicolumn{4}{c}{Real Data} \\ 
           & Small & Large-Sparse & Large   & Large-Dense & Cora ML & Cora    & Reddit      & Yelp       \\ 
\midrule
$N$& 10,000    & 100,000          & 100,000     & 100,000         & 2,995   & 19,793  & 232,965     & 716,847    \\ 
$M$
& 104,485   & 99,231           & 1,493,135   & 14,897,099      & 16,316  & 126,842 & 23,213,838  & 13,954,819 \\ 
$F$ & 100       & 100              & 100         & 100             & 2,879   & 8,710   & 602         & 300        \\ 
\bottomrule
\end{tabular}
\caption{\label{tab:dataset} Network data statistics:
{$N=$  no.\,nodes, $M=$  no.\,edges, $F=$no.\,features}}
\end{table}
}
}
\vspace{-1cm}
\paragraph{Experimental setup and Results}
\stratis{
\gdr{To assess whether L2G2G is}
a scalable alternative for the use of a GAE without having to sacrifice accuracy in downstream tasks, \gdr{we compare it against} 
the standard GAE~\cite{vgae}, GAE+L2G~\cite{local2global} and FastGAE~\cite{fastgae}.}
We train the models on each data set for 200 epochs, with learning rate 0.001 and the Adam optimizer \cite{adam}, {and} 
two layers in the GCN. The dimension of the first hidden layer
is 32 and the dimension of the last layer is 16.
\gdr{We} run each experiment 10 times with different random seeds for each model on each data set.
All the experiments were conducted on 
{a} V100 GPU.
We then compare the \gdr{methods using the} Area Under the Curve (AUC)  and the Average Precision (AP).
\gdr{Following \cite{local2global}, we}  
test our algorithm with fixed patch size 10.

Table~\ref{tab:patch10} shows that  L2G2G outperforms both FastGAE and GAE+{L2G} \stratis{on most experiments.}
{Having established the theoretical training speed gain of L2G2G, these results illustrate that L2G2G can} perform better than the GAE+{L2G}, as well as achieve comparable training speed. 
\AEEr{do we need to clarify with new pytorch stuff?}
Furthermore, in contrast to FastGAE we observe that the performance of L2G2G and of the GAE are very close to each other on the medium and large-scale data sets, indicating that L2G2G {does not lose much performance compared to the much slower but assumed more accurate classic GAE.} 
Furthermore, L2G2G even outperforms the GAE when the data set is large and dense, such as SBM-Large-dense and Reddit.

\begin{figure}[t]
    \centering
    \includegraphics[width=0.65\linewidth,trim={0.10cm 0.30cm 0.40cm 0.10cm},clip]{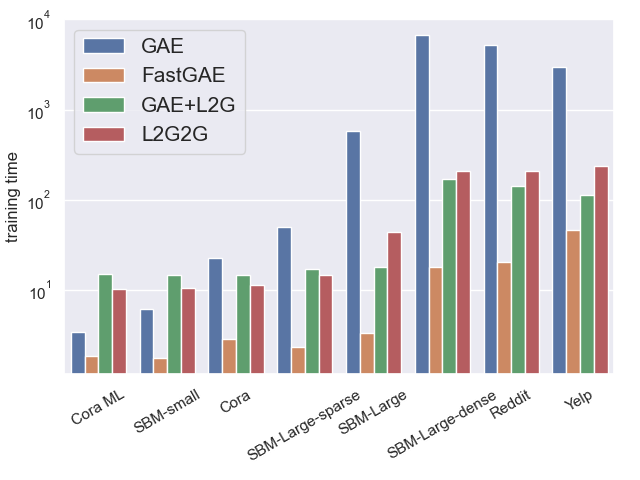}
		\caption{Training time of the baseline models(GAE, FastGAE and GAE+{L2G}) and L2G2G on benchmark data sets {(excluding partitioning time)}. Note that the y-axis is on a log-scale, and thus the faster methods are at least an order of magnitude faster. 
	}
    \label{fig:patch10 training time}
\end{figure}

\begin{table}[t]
\begin{tabular}{l@{\hskip 0.2cm}|@{\hskip 0.2cm} c@{\hskip 0.2cm}| @{\hskip 0.2cm}c@{\hskip 0.2cm}| @{\hskip 0.2cm}c@{\hskip 0.2cm}| @{\hskip 0.2cm}c}
\toprule
\multicolumn{5}{c}{\textbf{Average Performance On  Different Datasets (AUC in \%)}}                                                                                                                                                        \\ \midrule
                     & GAE                  &  FastGAE              & GAE+L2G              & L2G2G                 \\
                     \hline
Cora ml              & 95.95 ± 0.42         &  83.90 ± 1.10         & 90.25 ± 0.19         & \textbf{92.58 ± 0.35} \\
SBM-small            & 95.32 ± 0.18         &  76.34 ± 0.57         & 93.84 ± 0.14         & \underline{\textbf{95.39 ± 0.21}} \\
Cora                 & 96.07 ± 0.09         &  81.78 ± 0.76         & 90.59 ± 0.11         & \textbf{94.96 ± 0.26} \\ 
\hline
SBM-Large-sparse     & 94.88 ± 0.23         &  80.89 ± 0.84         & \text{94.73 ± 0.07}  & \underline{\textbf{95.02 ± 0.23}} \\
SBM-Large            & 86.84 ± 0.11         &  70.90 ± 1.29         & \text{84.60 ± 0.10}  & \textbf{86.62 ± 0.25} \\ 
SBM-Large-dense      & 64.07 ± 0.12         &  65.20 ± 0.94         & 65.45 ± 0.05         & \underline{\textbf{65.88 ± 0.03}} \\
\hline
Reddit               & 88.69 ± 0.57         &  79.99 ± 1.31         & \textbf{88.50 ± 0.23}& \text{88.37 ± 0.39}   \\
Yelp                 & 86.55 ± 0.28         &  73.79 ± 6.54         & \textbf{85.82 ± 0.17}& 84.01 ± 0.11          \\
\bottomrule
\end{tabular}

\begin{tabular}{l@{\hskip 0.2cm}|@{\hskip 0.2cm} c@{\hskip 0.2cm}| @{\hskip 0.2cm}c@{\hskip 0.2cm}| @{\hskip 0.2cm}c@{\hskip 0.2cm}| @{\hskip 0.2cm}c}
\toprule
\multicolumn{5}{c}{\textbf{Average Performance On  Different Datasets (AP in \%)}}                                                                                                                                                         \\ \midrule
                     & GAE                   & FastGAE              & GAE+L2G               & L2G2G                  
\\ \hline
Cora ml              & 95.37 ± 0.57          & 83.90 ± 1.10         & 90.57 ± 0.19          & \textbf{92.41 ± 0.39}  
\\
SBM-small            & 95.23 ± 0.12          & 76.34 ± 0.57         & 95.22 ± 0.11          & \underline{\textbf{95.71 ± 0.24}}  
\\
Cora                 & 95.76 ± 0.14          & 81.78 ± 0.76         & 90.50 ± 0.15          & \textbf{94.67 ± 0.29}  
\\
SBM-Large-sparse     & 95.26 ± 0.24          & 80.89 ± 0.84         & \underline{\textbf{95.88 ± 0.07}} & \text{95.44 ± 0.26}    
\\
SBM-Large            & 89.64 ± 0.21          & 70.90 ± 1.29         & \text{87.35 ± 0.11}   & \textbf{89.34 ± 0.33}  
\\
SBM-Large-dense      & 67.64 ± 0.30          & 65.20 ± 0.94         & 71.25 ± 0.06          & \underline{\textbf{72.08 ± 0.05}}  
\\
Reddit               & 88.16 ± 0.60          & 79.99 ± 1.31         & 88.40 ± 0.18   & \underline{\textbf{88.57 ± 0.40}}  
\\
Yelp                 & 86.73 ± 0.29          & 73.79 ± 6.54         & \textbf{85.26 ± 0.12} & 83.56 ± 0.11           
\\ \bottomrule
\end{tabular}

\caption{\label{tab:patch10}Experiments on different data sets with patch size 10. \gdr{Bold:} 
the best \gdr{among the} 
fast methods, \gdr{underlined:} 
the model outperforms the GAE.
}

\end{table}
Figure~\ref{fig:patch10 training time} shows a 
comparison of the training time of the models, 
as well as 
the changes of training speed as the data set size increases \gdr{(on the log scale)}. 
\stratis{It is worth mentioning that we are not accounting for} the run time of the graph clustering.
{The results show that}
the training speed of L2G2G and GAE+{L2G} are very close on both small and large scale datasets.
{Although} the gap between the training speed of L2G2G and that of GAE+{L2G} {increases for}
 large-scale data sets, L2G2G still achieves high training speed, {and is even not much slower that FastGAE while achieving much better performance.}
\gdr{In almost all cases,} L2G2G is faster than the 
\gdr{standard} GAE, 
\stratis{except for the two smaller datasets}. 
\gdr{Its} training time  is around an order of magnitude smaller per epoch for the larger models. \gdr{As an aside,} \stratis{
GAEs suffer from memory issues as they need to store very large matrices during the decoding step.}

\begin{figure}[ht!]
     \centering
     \begin{minipage}{0.3\textwidth}
     \centering
     {\footnotesize Cora ML}
     \includegraphics[width=\textwidth,trim={0.3cm 0.1cm 0.3cm 0.0cm},clip]{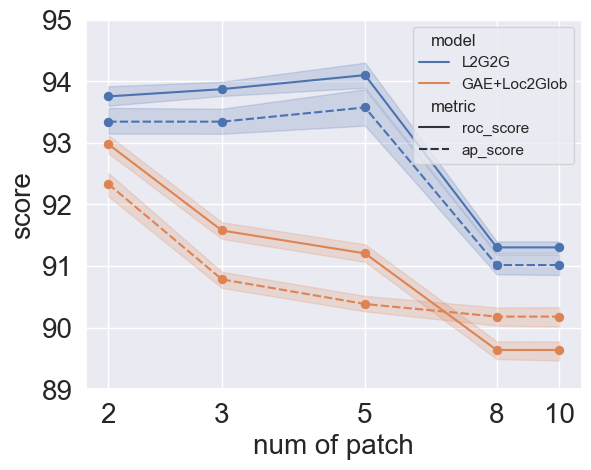}
     \end{minipage}
     \begin{minipage}{
     0.3\textwidth}
     \centering
     {\footnotesize Cora} 
     \includegraphics[width=\textwidth,trim={0.3cm 0.1cm 0.3cm 0.0cm},clip]{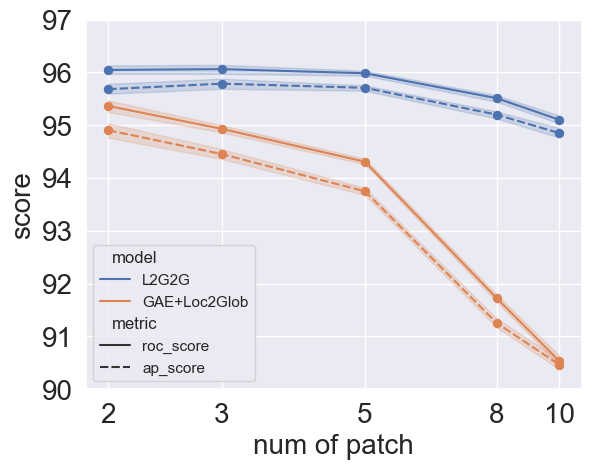}
     \end{minipage}
     \begin{minipage}{
     0.3\textwidth}
     \centering
     {\footnotesize  SBM-small} 
 
     \includegraphics[width=\textwidth,trim={0.3cm 0.1cm 0.3cm 0.0cm},clip]{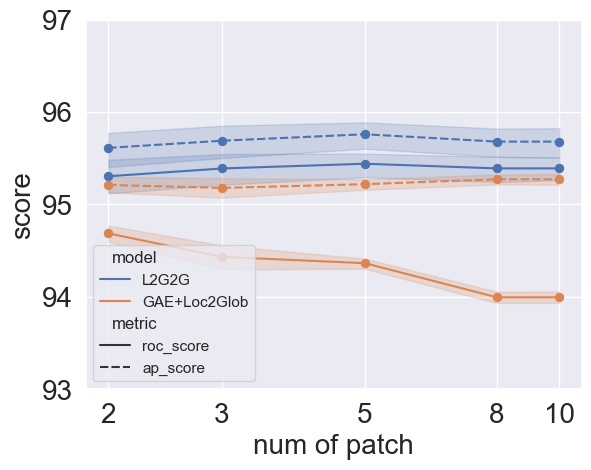}
     \end{minipage} 
    \\ 
     \vfill
     \begin{minipage}{0.3\textwidth}
     \centering
     {\footnotesize SBM-Large} 
 
     \includegraphics[width=\textwidth,trim={0.3cm 0.1cm 0.3cm 0.0cm},clip]{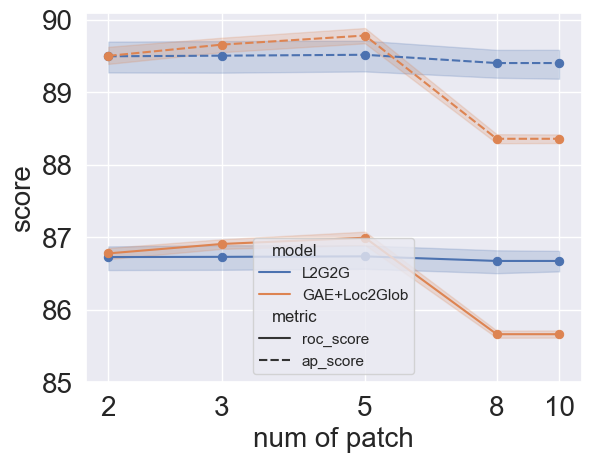}
     \end{minipage}
     \begin{minipage}{0.3\textwidth}
     \centering
     {\footnotesize Reddit} 
 
     \includegraphics[width=\textwidth,trim={0.3cm 0.1cm 0.3cm 0.0cm},clip]{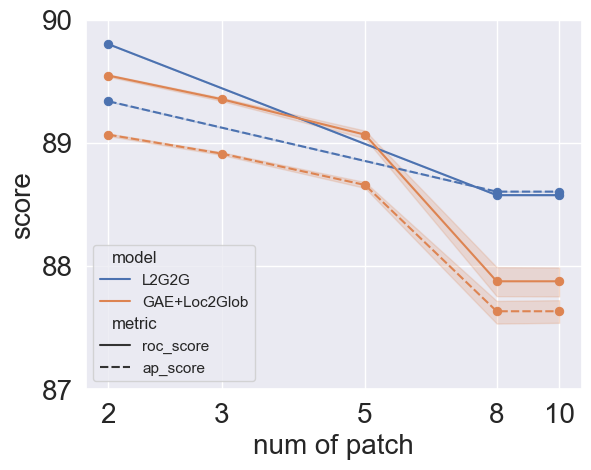}
     \end{minipage}
     \begin{minipage}{0.3\textwidth}
     \centering
     {\footnotesize  Yelp }
 
     \includegraphics[width=\textwidth,trim={0.3cm 0.1cm 0.3cm 0.0cm},clip]{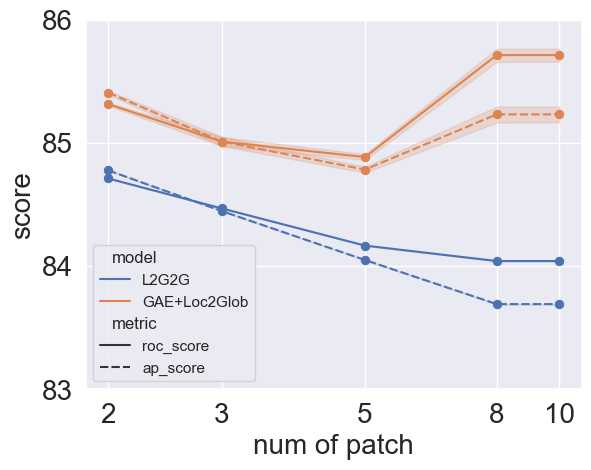}
     \end{minipage}
    \caption{Lineplots of the ROC score and accuracy of L2G2G and GAE+L2G, trained on each dataset, with different patch sizes. For each subplot, the blue lines represent the metrics  for L2G2G, while the orange ones represent those for  GAE+L2G. The shadows in each subplot indicate the standard deviations of each metric.
    }
    \label{fig:comparison performance}
\end{figure}

\begin{figure}[ht!]
\centering
        \begin{tabular}{@{}c@{}c@{}c@{}}
        & CPU & GPU \\ 
        \rotatebox{90}{\hspace{1.25cm}Cora}
       &
         \includegraphics[width=0.35\linewidth]{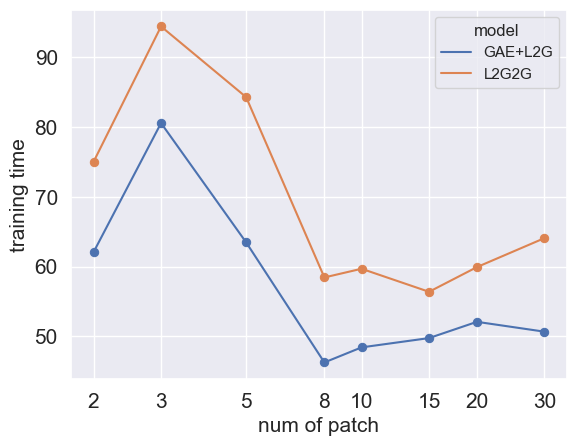} &
         \includegraphics[width=0.35\linewidth]{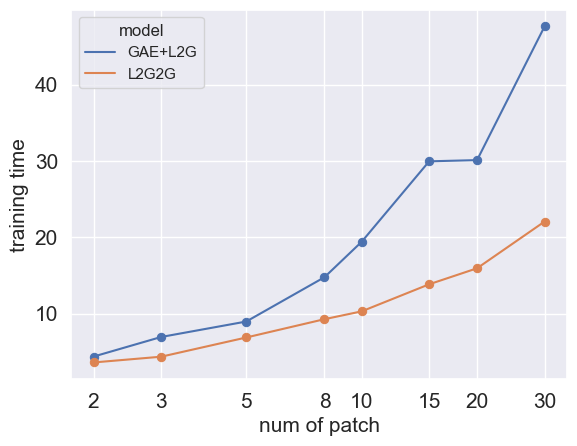}
         \\
        \rotatebox{90}{\hspace{1.25cm} 
        Yelp}
       &
         \includegraphics[width=0.35\linewidth]{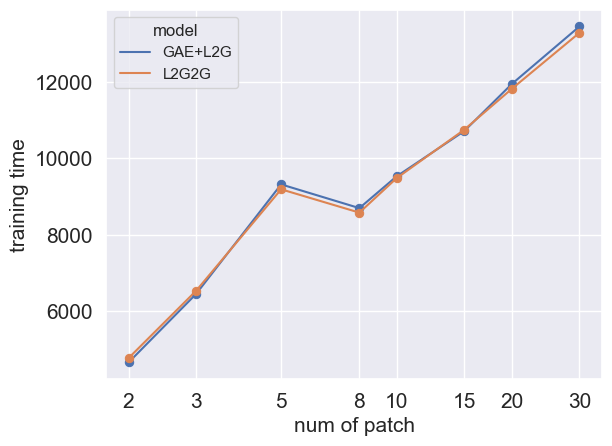} &
         \includegraphics[width=0.35\linewidth]{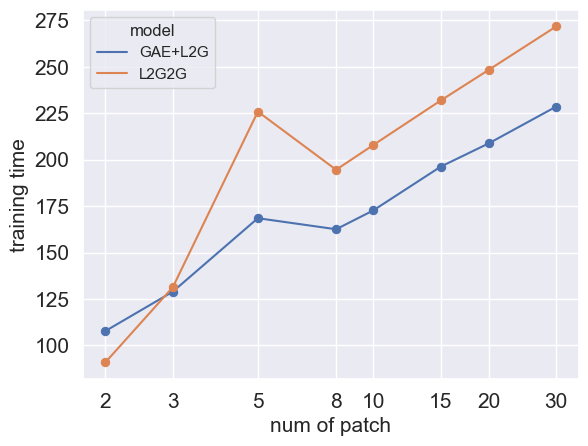}
        \end{tabular}
    \caption{Training time {(excluding partitioning)} of L2G2G and GAE+L2G 
on Cora 
({\bf Top})
and
     Yelp
({\bf Bottom}), while varying patch size
    with CPU results presented on the {\bf left} and GPU results presented on the {\bf right}. The x axis is shown in log scale. \
    }
    \label{fig:comparison time}
\end{figure}

\paragraph{Ablation Study}
\gdr{Here} we vary the number of patches, ranging from 2 to 10.
Figure~\ref{fig:comparison performance} shows the performance changes with different number of patches for each model on each data set. 
\gdr{W}hen the patch size increases, the performance of L2G2G decreases less than GAE+{L2G}. \stratis{This shows that updating the node embeddings dynamically during the training and keeping the local information with the agglomerating loss actually brings stability to \gdr{L2G2G}.
}
 \gdr{Moreover,}
we have explored the behaviour of training time for \gdr{L2G2G} when patch size increases from 2 to 30, on both a small (Cora) and a large (Yelp) dataset.
Figure~\ref{fig:comparison time} shows that on the small-scale data set Cora, the gap in training speed between L2G2G and GAE+{L2G} remains almost unchanged, while on Yelp, the gap between \lgg and GAE+{L2G} becomes smaller. However,
the construction of the overlapping patches in the Local2Global library 
{can} 
create patches that are much larger than ${N}/{k}$, 
potentially resulting in a large number of nodes in each patch. 
Hence, the training time in our tests increases with 
the number of patches.

Since all the computations in Local2Global library built by \cite{local2global} are carried out on the CPU, the GPU training can be slowed down by the memory swap between CPU and GPU. Thus, to further explore the behaviour of our algorithm when the number of patches increases, we ran the test on both CPU and GPU. The results are given by Figure~\ref{fig:comparison time}.
{This plot illustrates that the GPU training time of L2G2G increases moderately with increasing patch size, mimicking the behaviour of GAE+L2G. In contrast, the CPU training time for the smaller data set (Cora) decreases with increasing patch size. The larger but much sparser Yelp data set may not lend itself naturally to a partition into overlapping patches. 
} 
{\gdr{Summarising,} L2G2G performs better than the baseline models across most settings,  \gdr{while} sacrificing a tolerable amount of training speed.

\section{Conclusion {and Future Work}} \label{sec:conclusion} 
{In this paper, we have introduced L2G2G, a fast yet accurate method for obtaining node embeddings for large-scale  networks. In our experiments, L2G2G outperforms FastGAE and GAE+{L2G}, \stratis{while the amount of training speed sacrificed is tolerable}
We also find that L2G2G is not as sensitive to patch size change as GAE+{L2G}.} 

{Future work  will investigate embedding the synchronization step in the network  instead of performing the Local2Global algorithm to align the local embeddings. This change would potentially avoid matrix inversion, speeding up the calculations. We shall also investigate the performance on stochastic block models with more heterogeneity.} 

{To improve accuracy,}
one could add a small number of between--patch losses into the \lgg loss function, {to}   
account for edges which do not fall within a patch. The additional complexity of this {change} would be
relatively limited {when} restricting the number of between--patches included. 
Additionally, the Local2Global library 
from \cite{local2} is implemented on CPU, {losing speed}
%
{due to} moving memory between the CPU and the GPU. 
{W}e will investigate re-implementing the Local2Global algorithm on a GPU.

%
%
%
\bibliographystyle{spmpsci}
%
\bibliography{sample}








\end{document}